\documentclass{article}
\usepackage{graphicx} 
\usepackage[numbers]{natbib}
\usepackage{booktabs}
\usepackage{xcolor}
\usepackage{multirow}
\usepackage{url}
\usepackage{graphicx}
\usepackage{spverbatim}
\usepackage{amsmath}
\usepackage{float}
\setcitestyle{square}
\usepackage{authblk}
\usepackage{geometry}
\usepackage{titlesec}
\usepackage{newtxtext}
\usepackage{newtxmath}
\usepackage{sectsty}
\usepackage{hyperref}
\usepackage[misc]{ifsym}
\usepackage{authblk}
\usepackage{booktabs}
\usepackage{pdfpages}

\title{Development and Validation of the Provider Documentation Summarization Quality Instrument for Large Language Models}
\author{Emma Croxford$^{2}$, Yanjun Gao PhD$^{5}$, Nicholas Pellegrino$^{7}$, Karen K. Wong MD$^{7,8}$, \\Graham Wills PhD$^{8}$, Elliot First$^{7}$, Miranda Schnier$^{7}$, Kyle Burton MD$^{8}$, Cris G. Ebby MD, MS$^{4}$, Jillian Gorskic MD$^{8}$, Matthew Kalscheur MD$^{1,8}$, Samy Khalil MD$^{8}$, Marie Pisani BS$^{1}$, Tyler Rubeor MD$^{8}$, Peter Stetson, MD MA$^{6}$, Frank Liao PhD$^{3, 8}$, Cherodeep Goswami$^{8}$, Brian Patterson MD$^{3,8}$, Majid Afshar MD\href{mailto:mafshar@medicine.wisc.edu}{\Letter}$^{8,}$} 

\date{}

\affil[1]{Department of Medicine, University of Wisconsin, Madison}
\affil[2]{Department of Biostatistics and Medical Informatics, University of Wisconsin, Madison}
\affil[3]{BerbeeWalsh Department of Emergency Medicine, University of Wisconsin, Madison}
\affil[4]{Department of Pediatrics, University of Wisconsin, Madison}
\affil[5]{Department of Biomedical Informatics, University of Colorado - Anschutz Medical}
\affil[6]{Memorial Sloan Kettering Cancer Center, New York, NY}
\affil[7]{Epic Systems, Verona, WI}
\affil[8]{UW Health, Madison, WI\vspace{-1ex}\vspace{-1ex}}

\begin{document}

\maketitle

\section{Abstract}

Background: As Large Language Models (LLMs) are integrated into electronic health record (EHR) workflows, validated instruments are essential to evaluate their performance before implementation and as models and documentation practices evolve. Existing instruments for provider documentation quality are often unsuitable for the complexities of LLM-generated text and lack validation on real-world data. The Provider Documentation Summarization Quality Instrument (PDSQI-9) was developed to evaluate LLM-generated clinical summaries. This study aimed to validate the PDSQI-9 across key aspects of construct validity.

Methods: Multi-document summaries were generated from real-world EHR data across multiple specialties using several LLMs (GPT-4o, Mixtral 8x7b, and Llama 3-8b). Validation included Pearson correlation analyses for substantive validity, factor analysis and Cronbach’s $\alpha$ for structural validity, inter-rater reliability (ICC and Krippendorff’s $\alpha$) for generalizability, a semi-Delphi process for content validity, and comparisons of high- versus low-quality summaries for discriminant validity. Raters underwent standardized training to ensure consistent application of the instrument.

Results: Seven physician raters evaluated 779 summaries and answered 8,329 questions, achieving over 80\% power for inter-rater reliability. The PDSQI-9 demonstrated strong internal consistency (Cronbach’s $\alpha$ = 0.879; 95\% CI: 0.867–0.891) and high inter-rater reliability (ICC = 0.867; 95\% CI: 0.867–0.868), supporting structural validity and generalizability. Factor analysis identified a 4-factor model explaining 58\% of the variance, representing organization, clarity, accuracy, and utility. Substantive validity was supported by correlations between note length and scores for \textit{Succinct} ($\rho$ = -0.200, p = 0.029) and \textit{Organized} ($\rho$ = -0.190, p = 0.037). The semi-Delphi process ensured clinically relevant attributes, and discriminant validity distinguished high- from low-quality summaries ($p < 0.001$).

Conclusions: The PDSQI-9 demonstrates robust construct validity, supporting its use in clinical practice to evaluate LLM-generated summaries and facilitate safer, more effective integration of LLMs into healthcare workflows.

\section{Introduction}

The volume of notes in the electronic health record (EHR) has increased in the past decade, exacerbating the burden on providers and highlighting the growing difficulty of the unassisted “chart biopsy” task.  One in five patients arrives at the hospital with a chart comparable to the size of "\textit{Moby Dick}" (206,000 words). ~\cite{Patterson_Hekman_Liao_Hamedani_Shah_Afshar_2024} While the EHR's centralized storage of medical notes is beneficial, and studies have shown that access to prior records improves diagnostic accuracy, the growing volume of data presents a significant challenge. ~\cite{Institute_America_2000} The tension between the EHR's role as a documentation repository and its function as a tool for retrieving actionable information has become increasingly unmanageable without additional filtering and summarization tools. ~\cite{Embi_Weir_Efthimiadis_Thielke_Hedeen_Hammond_2013}     

In the field of clinical Natural Language Generation (NLG), multi-document summarization has emerged as an important task to address the challenge of note bloat and reduce the cognitive burden on healthcare providers. As Large Language Models (LLMs) continue to advance NLG capabilities, they offer a promising alternative for summarizing clinical documentation and alleviating the time-intensive nature of human-authored summaries. However, this capability also introduces challenges, including performance degradation with chronological errors or missed details. ~\cite{Liu_Lin_Hewitt_Paranjape_Bevilacqua_Petroni_Liang_2024} Despite their potential, the rapid advancements of LLMs have outpaced the development of robust evaluation instruments to assess the quality of their outputs. 

There remains a paucity of evidence on human evaluation instruments designed for LLM summarizations developed from real-world, multi-document EHR data and supported by psychometric validation. ~\cite{Bedi} Tam et al.'s systematic review of 142 studies on human evaluation methodologies for LLMs in healthcare highlighted significant gaps, including the lack of sample size calculations, insufficient details on the number of evaluators, and inadequate evaluator training. ~\cite{Tam_Sivarajkumar_Kapoor_Stolyar_Polanska_McCarthy_Osterhoudt_Wu_Visweswaran_Fu_etal._2024} While their paper provided areas for improvement, the suggested strategies were primarily based on patterns observed in the literature rather than being grounded in statistical frameworks. There is a significant gap in evaluation methodologies designed to address the complexities of healthcare applications and the unique challenges posed by LLMs. 

Existing evaluative instruments for provider notes are designed primarily for provider-authored documentation. One widely adopted instrument, which has also been applied to AI-generated notes ~\cite{Kernberg_Gold_Mohan_2024, Owens_Wilda_Grifka_Westendorp_Fletcher_2024, Tierney_Gayre_Hoberman_Mattern_Ballesca_Kipnis_Liu_Lee_2024}, is the Physician Documentation Quality Instrument (PDQI). The instrument has demonstrated strong reliability, achieving high inter-rater reliability and internal consistency. ~\cite{Stetson_Bakken_Wrenn_Siegler_2012} Although the PDQI-9 tool is validated and reliable for evaluating provider-authored notes, it was not designed to address the unique challenges posed by LLM summarization of notes. Summaries generated by LLMs must be assessed for additional factors such as relevancy, hallucinations, omissions, and factual accuracy, areas where LLMs have demonstrated limitations. ~\cite{Zhao_Zhou_Li_Tang_Wang_Hou_Min_Zhang_Zhang_Dong_etal._2023} To address these gaps, this study introduces the Provider Documentation Summarization Quality Instrument (PDSQI-9), an LLM-centric adaptation of the PDQI-9. The PDSQI-9 is specifically designed to evaluate LLM-generated summaries, developed and validated on real-world EHR data, and rigorously tested for psychometric properties with adequate statistical power.

\section{Methods}

\subsection{Study Design, Setting, and Data}

The corpus of notes was designed for multi-document summarization and evaluation using inpatient and outpatient encounters from the University of Wisconsin Hospitals and Clinics (UW Health) in Wisconsin and Illinois between March 22, 2023 and December 13, 2023. The evaluation was conducted from the perspective of the provider during their initial \textit{office visit} with the patient ("index encounter"), representing a real-world clinic appointment where the provider benefits from a summary of the patient's prior encounters with outside providers. Other inclusion and exclusion criteria were the following: (1) patient was  alive at time of index encounter with provider; (2) patient had at least one encounter in 2023; and (3) excluded psychiatry notes. The corpus was further filtered to patients with 3 or more encounters before the index encounter, to provide a multi-document occurrence and fit the context window of many large language models. The resultant corpus consisted of 2,123 patients, with 554 having 3 encounters, 389 having 4 encounters, and 1,180 having 5 encounters. The derived dataset, consisting of encounters with concatenated provider notes leading to the index encounter of interest, was built as a random sample from the corpus with stratification across 11 specialties (25\% from gynecology, urgent care, neurosurgery, neurology, or urology; 40\% from dermatology, surgery, orthopedics, or ophthalmology; 35\% from family medicine or internal medicine) The sample size needed for evaluation was determined a priori (see sample size estimation) and included additional examples for pilot testing by the instrument developers and training by the raters. The final dataset was 200 unique patients and their encounters and 22.5\%, 22\%, and 55.5\% had 3, 4, and 5 notes per patient summarized prior to the index encounter. This study was approved by the University Wisconsin-Madison Institutional Review Board and qualified as exempt human subjects research.

\subsection{Development of Provider Document Summarization Quality Instrument (PDSQI)-9}

The instrument development process employed a semi-Delphi methodology, an iterative consensus-driven approach commonly used for gathering expert opinions and refining complex frameworks. ~\cite{Turoff_Linstone} The semi-Delphi process consisted of three iterative rounds, each involving nine stakeholders with diverse expertise: three physicians who were also clinical informaticists with specialized knowledge in human factors design and natural language processing (MA, BP, KW); two software developers with experience in generative AI (NP, EF); one quality improvement specialist (MS); two data scientists (EC, GW); and one computer scientist with expertise in computational linguistics (YG). 

\textit{Round 1 - Literature Review and Domain Identification:}
The panel reviewed existing literature and methodologies for evaluating clinical text summarizations. The PDQI-9 was selected as the benchmark due to its demonstrated validity, interpretability, and applicability to evaluating physician clinical documentation. The panel then identified key domains essential for high-quality multi-document summarization, as well as dimensions where LLMs are known to underperform, such as hallucinations, omissions, and relevancy.

The identified dimensions were mapped to existing PDQI-9 attributes where feasible, with modifications to improve applicability of clarity and relevance attributes. Two PDQI-9 attributes, \textit{Up-to-Date} and \textit{Consistent}, were removed as their conceptual scope was adequately captured by modifications to other attributes. Two new attributes were added to address concerns in LLM-generated summaries: use of stigmatizing language and inclusion of citations linking facts in the summary to the original documentation. Additionally, the panel placed a particular focus on the vulnerability of each domain to hallucinations, defined as falsifications or fabrications. 

\textit{Round 2 - Attribute Refinement and Mapping to Likert Scales:}
The instrument definitions for each attribute were refined, and detailed instructions were developed for scoring on a five-point Likert scale. The panel iteratively revised attribute definitions to ensure clarity and usability. Special emphasis was placed on designing attribute definitions that captured the nuances of clinical text summarization, including factors such as relevancy, factual accuracy, and faithfulness to the source documentation.

\textit{Round 3 - Pilot Testing and Consensus Adjudication:}
Pilot testing was performed with three senior physicians (MA, KW, BP) from different specialties to evaluate the usability and clarity of the attributes and scoring instructions. Feedback from these testers was incorporated iteratively, with the adjudication of disagreements conducted by the expert panel to achieve consensus. Pilot testing continued until all testers agreed on the final instrument definitions and scoring instructions, ensuring content validity of the instrument through the semi-Delphi process. The final instrument is in Appendix C.

\begin{table} [H]
    \centering
    \begin{tabular}{p{0.2\linewidth} | p{0.5\linewidth} | p{0.2\linewidth}} \toprule
        \textbf{PDSQI-9 Attribute} & \textbf{Definition} & \textbf{Relevant Domain(s)}\\ \midrule
        Accurate & 
        The summary is true and free of incorrect information & 
            Extraction ~\cite{Sai_Mohankumar_Khapra_2023, Zhao_Zhou_Li_Tang_Wang_Hou_Min_Zhang_Zhang_Dong_etal._2023}, 
            Faithfulness ~\cite{Cai_Liu_Bajracharya_Sills_Kapoor_Liu_Berlowitz_Levy_Pradhan_Yu_2022, Adams_Zucker_Elhadad_2023}, 
            Recall ~\cite{Singhal_Azizi_Tu_Mahdavi_Wei_Chung_Scales_Tanwani_Cole-Lewis_Pfohl_etal._2023}, 
            Falsification/Fabrication ~\cite{Umapathi_Pal_Sankarasubbu_2023} 
            \\ \midrule
        Cited & 
        The summary includes citations that are present and appropriate &
            Rationale ~\cite{Singhal_Azizi_Tu_Mahdavi_Wei_Chung_Scales_Tanwani_Cole-Lewis_Pfohl_etal._2023} 
            \\ \midrule
        Comprehensible & 
        The summary is clear, without ambiguity or sections that are difficult to understand & 
            Coherence ~\cite{Wallace_Saha_Soboczenski_Marshall_2020}, 
            Fluency ~\cite{Otmakhova_Verspoor_Baldwin_Lau_2022} 
            \\ \midrule
        Organized & 
        The summary is well-formed and structured in a way that helps the reader understand the patient’s clinical course & 
            Structure ~\cite{Cohan_Goharian},
            Up-to-Date ~\cite{Stetson_Bakken_Wrenn_Siegler_2012},
            Currency ~\cite{Tam_Sivarajkumar_Kapoor_Stolyar_Polanska_McCarthy_Osterhoudt_Wu_Visweswaran_Fu_etal._2024}
            \\ \midrule
        Succinct & 
        The summary is brief, to the point, and without redundancy &
            Specificity ~\cite{Croxford_Gao_Patterson_To_Tesch_Dligach_Mayampurath_Churpek_Afshar_2024}, 
            Syntax ~\cite{Otmakhova_Verspoor_Baldwin_Lau_2022}, 
            Semantics ~\cite{Yadav_Gupta_Abacha_Demner-Fushman_2021}
            \\ \midrule
        Stigmatizing & 
        The summary is free of stigmatizing language & 
            Bias ~\cite{Singhal_Azizi_Tu_Mahdavi_Wei_Chung_Scales_Tanwani_Cole-Lewis_Pfohl_etal._2023}, 
            Harm ~\cite{Singhal_Azizi_Tu_Mahdavi_Wei_Chung_Scales_Tanwani_Cole-Lewis_Pfohl_etal._2023} 
            \\ \midrule
        Synthesized & 
        The summary reflects an understanding of the patient’s status and ability to develop a plan of care & 
            Abstraction ~\cite{Sai_Mohankumar_Khapra_2023, Zhao_Zhou_Li_Tang_Wang_Hou_Min_Zhang_Zhang_Dong_etal._2023}, 
            Reasoning ~\cite{Singhal_Azizi_Tu_Mahdavi_Wei_Chung_Scales_Tanwani_Cole-Lewis_Pfohl_etal._2023} ,
            Consistency ~\cite{Yadav_Gupta_Abacha_Demner-Fushman_2021, Guo_Qiu_Wang_Cohen_2022}
            \\ \midrule
        Thorough & 
        The summary should thoroughly cover all pertinent patient issues & 
            Omission ~\cite{Abacha_Yim_Michalopoulos_Lin_2023},
            Comprehensiveness ~\cite{Tam_Sivarajkumar_Kapoor_Stolyar_Polanska_McCarthy_Osterhoudt_Wu_Visweswaran_Fu_etal._2024}
            \\ \midrule
        Useful & 
        The summary is relevant, providing valuable information and/or analysis & 
            Plausibility ~\cite{Croxford_Gao_Patterson_To_Tesch_Dligach_Mayampurath_Churpek_Afshar_2024}, 
            Relevancy ~\cite{Wallace_Saha_Soboczenski_Marshall_2020}
            \\ \bottomrule
    \end{tabular}
    \caption{\textbf{PDSQI-9 Attributes, Definitions, and Relevant Domains} A table outlining the 9 attributes of our Provider Documentation Summarization Quality Instrument. Each attribute is accompanied by a description of the attribute that was part of the instrument provided to evaluators. Additionally, the relevant evaluation domains that it are associated with the concept behind a particular attribute is provided with references.}
    \label{tab:domains}
\end{table}

The final instrument includes nine attributes: \textit{Cited}, \textit{Accurate}, \textit{Thorough}, \textit{Useful}, \textit{Organized}, \textit{Comprehensible}, \textit{Succinct}, \textit{Synthesized}, and \textit{Stigmatizing} (Table ~\ref{tab:domains}). While the original PDQI-9 tool used the same 5-point Likert scale for every attribute, our adapted version incorporates a combination of 5-point Likert scales and binary scales tailored to the specific requirements of each attribute. The PDSQI-9 was developed and managed using REDCap.  ~\cite{Harris_Taylor_Thielke_Payne_Gonzalez_Conde_2008, Harris_Taylor_Minor_Elliott_Fernandez_O’Neal_McLeod_Delacqua_Delacqua_Kirby_etal._2019} The REDCap instrument and data dictionary are available at the following GitLab repository: https://git.doit.wisc.edu/smph-public/dom/uw-icu-data-science-lab-public/pdsqi-9.

\subsection{LLM Summarizations for Validating the PDSQI-9}

To generate summaries of varying quality, we employed different prompts across different LLMs to summarize notes for each patient encounter leading up to the index encounter. The LLMs utilized in this study included OpenAI's GPT-4o ~\cite{OpenAI_Achiam_Adler_Agarwal_Ahmad_Akkaya_Aleman_Almeida_Altenschmidt_Altman_etal._2024}, Mixtral 8x7B ~\cite{Jiang_Sablayrolles_Roux_Mensch_Savary_Bamford_Chaplot_Casas_Hanna_Bressand_etal._2024}, and Meta's Llama 3-8B ~\cite{Grattafiori_Dubey_Jauhri_Pandey_Kadian_Al-Dahle_Letman_Mathur_Schelten_Vaughan_etal._2024} (Table \ref{tab:llms}). GPT-4 operates within the secure environment of the health system's HIPAA-compliant Azure cloud. No PHI was transmitted, stored, or used by OpenAI for model training or human review. All interactions with proprietary closed-source LLMs were fully compliant with HIPAA regulations, maintaining the confidentiality of patient data. The open-source LLMs, Mixtral 8x7B and Llama-3-8B, were downloaded from HuggingFace ~\cite{HuggingFace._2024} to HIPAA-compliant, on-premise servers.

\begin{table} [h]
    \centering
    \begin{tabular}{lccccc} \toprule
        \textbf{Model} & \textbf{Parameters} & \textbf{Context Window} & \textbf{Temperature} & \textbf{Top P} & \textbf{Max New Tokens} \\ \midrule
         GPT-4o & -- & 128,000 & 0.05 & 0.05 & -- \\ \midrule
         Mixtral 8x7b & 7b & 32,000 & 1.0 & 1.0 & 1000 \\ \midrule
         Llama 3-8b & 8b & 8,000 & 1.0 & 1.0 & 1000 \\ 
         \bottomrule
    \end{tabular}
    \caption{\textbf{Large Language Model Parameter Settings} For every model, we present the number of parameters and context window length reported in each model's technical specifications. The temperature, top p, and max new tokens settings are also reported here. Any additional settings were left to the defaults unless otherwise specified.}
    \label{tab:llms}
\end{table}

We used four strategies for engineering the evaluation prompts: minimizing perplexity, in-context examples, chain-of-thought reasoning, and self-consistency. The prompt for each LLM included a persona with the following instruction: "You are an expert doctor. Your task is to write a summary for a specialty of [\textit{target specialty}], after reviewing a set of notes about a patient." To generate lower-quality summaries, additional variations of the prompt removed instructions or encouraged the inclusion of false information. The persona and instruction were followed by two chains of thought: Rules and Anti-Rules, delineating positive and negative summarization steps. The Rules targeted specific attributes in the PDQSI-9 to generate high-quality summaries. Anti-Rules introduced intentional errors (e.g., hallucinations, omissions) to include mistakes. For each patient, the LLM was provided a randomized subset of Rules and Anti-Rules, ensuring heterogeneity in the generated summaries. The open-source models, Mixtral 8x7b and Llama 3-8b, were tasked with producing the lowest-quality summaries and were exclusively provided Anti-Rules as instructions. This approach aimed to provide a wide distribution of PDSQI-9 scores and to allow for discriminant validity testing. The prompts are available at https://git.doit.wisc.edu/smph-public/dom/uw-icu-data-science-lab-public/pdsqi-9. The final corpus had 100 summaries generated by GPT-4o, 50 by Mixtral 8x7b, and 50 by Llama 3-8b.

\subsection{Sample Size Estimation and Rater Training}
Sample size calculations were performed assuming a minimum of five raters and an even score distribution across a 5-point Likert scale. To achieve a desired statistical power of 80\% with a precision of 0.1, each rater was required to complete at least 84 evaluations. ~\cite{Rotondi_2018} 

Five junior physician raters with 1 to 5 years of post-graduate experience were recruited (MP, KB, CE, SK, TR). Additionally, two senior physician raters (JG, MK), each with at least 10 years of post-graduate experience, were recruited to complete a subset of evaluations for further validation. To standardize evaluation criteria and scoring, a group of three senior physician trainers (MA, KW, BP) conducted evaluations on three exemplar cases. These cases were subsequently used as reference materials for all raters during a live training session. Given the varying levels of expertise among the raters, all were provided with the Center for Health Care Strategies’ documentation on identifying bias and stigmatizing language in EHRs. ~\cite{Canonico} Following the training session, raters were encouraged to pose questions or highlight disagreements during subsequent practice days. After the training period, all raters independently completed three additional example cases. Agreement among the raters was established before proceeding with independent evaluations of the full dataset.

\subsection{Analysis Plan and Validation}

Baseline characteristics of the corpus notes and evaluators were analyzed. Distributions of evaluative scores for each attribute were visualized using density ridge plots. Token counts were derived using the Llama 3-8b tokenizer.

The PDSQI-9 was evaluated through multiple metrics to assess its validity and reliability, informed by Messick's Framework of validity. ~\cite{Messick} To examine the theoretical alignment of the PDSQI-9, Pearson correlation coefficients were calculated to evaluate relationships between input characteristics (e.g., note length) and attribute scores (e.g., \textit{Succinct} and \textit{Organized}). This tested whether the instrument captured expected relationships consistent with the theoretical underpinnings of summarization challenges and provided substantive validity. For generalizability, inter-rater reliability was assessed using the Intraclass Correlation Coefficient (ICC) and Krippendorf's ~\cite{Krippendorff_2018} $\alpha$, ensuring the instrument produced consistent results across evaluators with varying levels of expertise. ICC is derived from an analysis of variance (ANOVA) and has several forms tailored to specific use cases. ~\cite{Fisher_1992} In this study, a two-way mixed-effects model was used for consistency among multiple raters, specifically ICC(3,k). ~\cite{Koo_Li_2016}  Unlike ICC, which is a variance-based measure, Krippendorff's $\alpha$ was calculated based on observed disagreement among raters and adjusted for chance agreement. The performance of junior physician evaluators was tested against senior physician evaluators (JG, MK) using the Wilcoxon signed-rank test, to assess differences in median scores between the two groups. To assess the PDSQI-9's discriminative validity, a Mann-Whitney U test was performed between the lowest and highest quality summaries. The summaries generated by GPT-4o with error-free prompts were considered the highest quality, while those generated by Llama 3-8b and Mixtral 8x7b with error-prone prompts were considered the lowest quality. 

For structural validity, internal consistency was measured using Cronbach’s $\alpha$  ~\cite{Cronbach_1951}, which evaluates whether the instrument items reliably measure the same underlying construct. Confirmatory factor analysis was conducted to identify latent structures underlying the survey attributes and to evaluate alignment with theoretical constructs. Factor loadings were used to assess variance within the instrument. A four-factor model was selected based on eigenvalues, the scree plot, and model fit indices (Appendix B).

Cronbach's $\alpha$, ICC, and Krippendorff's $\alpha$ produced coefficients ranging between 0 and 1, where higher values indicate greater reliability or agreement. 95\% Confidence Intervals (CI) were provided for all coefficients and calculated using the Feldt procedure ~\cite{Feldt_Woodruff_Salih_1987}, Shrout \& Fleiss procedure ~\cite{Shrout_Fleiss}, and bootstrap procedure respectively. Analyses were performed using Python (version 3.11) \cite{transformers, nltk} and R Studio (version 4.3) ~\cite{smplot, ggplot, psych, kripp, icc}.

\section{Results}

Seven physician raters evaluated 779 summaries and scored 8,329 PDSQI-9 items to achieve greater than 80\% power for examining inter-rater reliability. No difference was observed in the scores by rater expertise when comparing junior and senior physicians (n = 48 summarizations; p-value = 0.187). The median time required for the junior physicians to complete a single evaluation, including reading the provider notes and the LLM-generated summary, was 10.93 minutes (IQR: 7.92–14.98). Senior physician raters completed evaluations with a median time of 9.82 minutes (IQR: 6.28–13.75) (Appendix A).

The provider notes, concatenated into a single input for each patient, had a median word count of 2,971 (IQR: 2,179–3,371) and a median token count of 5,101 (IQR: 3,614–7,464). The provider note types included notes from 20 specialties (medicine, family medicine, orthopedics, ophthalmology, emergency medicine, surgery, dermatology, urgent care, urology, neurology, gynecology, psychiatry, anesthesiology, neurosurgery, somnology, pediatrics, audiology, and radiology). The LLM-generated summaries of the provider notes had a median length of 328.5 words (IQR: 205.8–519.8) and 452.5 tokens (IQR: 313.5–749.5). A modest positive correlation was observed between the input text's length and the generated summaries' length ($\rho$ = 0.221 with p-value = 0.002). 

\begin{figure}[H]
    \centering
    \includegraphics[width=0.6\textwidth]{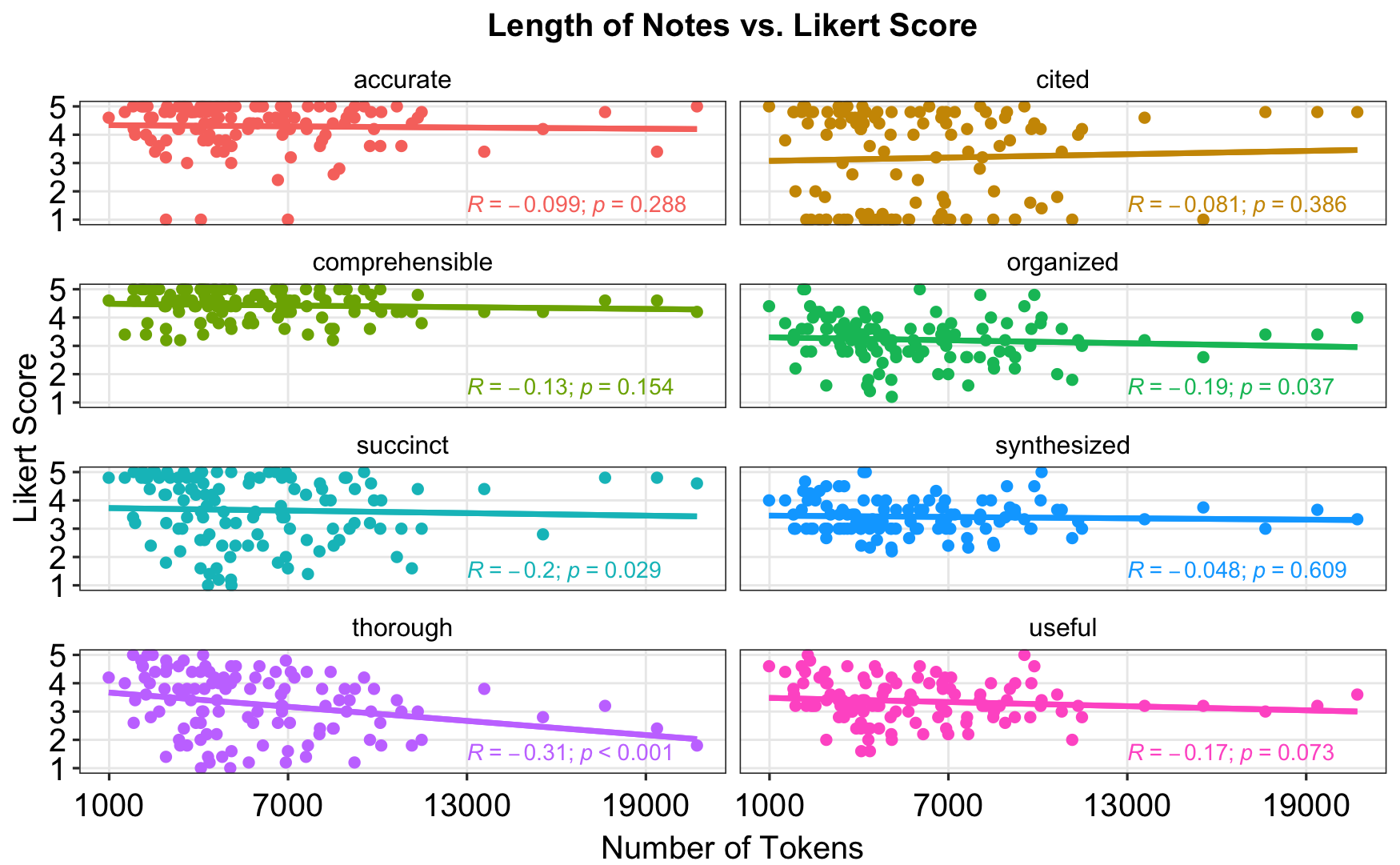}
    \caption{\textbf{Length of Patient Notes vs. Mean Evaluator Score} Scatter plots of the mean score among evaluators compared to the token length of the patient notes provided for summarization. Trend lines are superimposed on the plot along with the Spearman $\rho$ (denoted as R) coefficient and p-value (denoted as p). Each plot corresponds to a single attribute from the PDSQI-9 instrument.}
    \label{fig:length_score}
\end{figure}

\begin{figure}[H]
    \centering
    \includegraphics[width=0.6\textwidth]{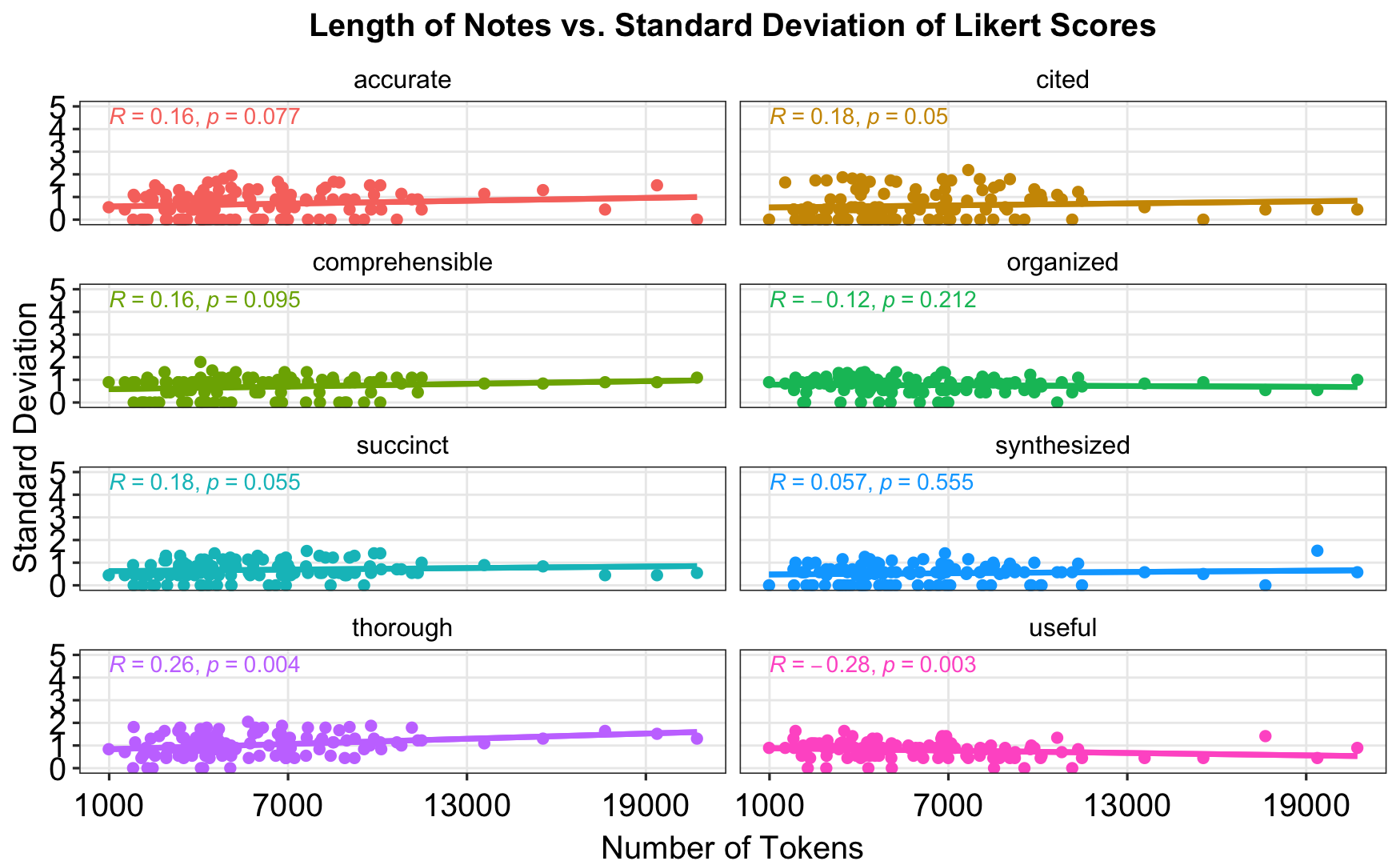}
    \caption{\textbf{Length of Patient Notes vs. Standard Deviation of Evaluator Scores} Scatter plots of the score standard deviation among evaluators compared to the token length of the patient notes provided for summarization. Trend lines are superimposed on the plot along with the Spearman $\rho$ (denoted as R) coefficient and its p-value (denoted as p). Each plot corresponds to a single attribute from the PDSQI-9 instrument.}
    \label{fig:length_sd}
\end{figure}

Figure ~\ref{fig:length_score} illustrates the average scores for each attribute of a summary, as evaluated by our raters, in relation to the length of the notes being summarized. As the length of the notes increased, the quality of the generated summaries was rated lower in the attributes of \textit{Organized} ($\rho$ = -0.190 with p-value = 0.037), \textit{Succinct} ($\rho$ = -0.200 with p-value = 0.029), and \textit{Thorough} ($\rho$ = -0.31 with p-value $<$ 0.001). Additionally, the variance in scores among the raters increased with longer note lengths for the attributes of \textit{Thorough} ($\rho$ = 0.26 with p-value = 0.004) and \textit{Useful} ($\rho$ = -0.28 with p-value = 0.003) (Figure ~\ref{fig:length_sd}).

\begin{figure}[H]
    \centering
    \includegraphics[width=0.7\textwidth]{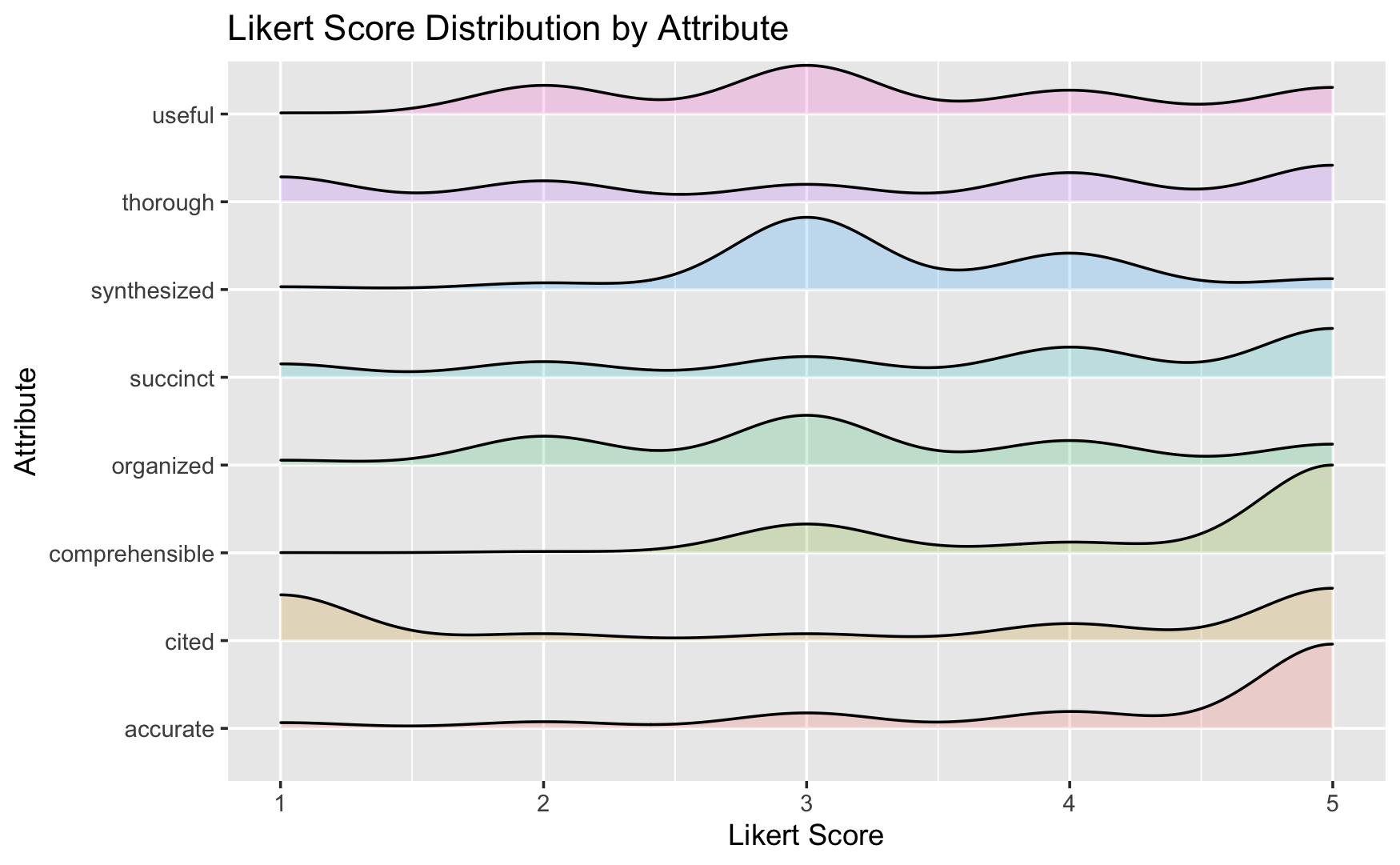}
    \caption{\textbf{Likert Score Distributions by Attribute} A density ridge across the 5 Likert scale points that evaluators used to score each attribute from the PDSQI-9 instrument. Each attribute is provided separately and identified on the y-axis. Distributions include the scores from every evaluator on every unique patient summary reported in this study.}
    \label{fig:dist}
\end{figure}

Figure ~\ref{fig:dist} illustrates the distribution of evaluative scores for each attribute of the PDQSI-9. As intended, the scores spanned the entire range of the PDQSI-9 for nearly all attributes. The only exception was the \textit{Comprehensible} attribute, where no summary received a score of "1" on the Likert scale. This was attributed to the inherent quality of the LLMs used and the challenges encountered in jailbreaking them to generate incomprehensible outputs. The attributes \textit{Succinct} and \textit{Thorough} exhibited the smoothest distributions in scoring. The median score and IQR for each attribute was 3.0 (3.0-4.0) for \textit{Useful}, 4.0 (2.0-5.0) for \textit{Thorough}, 3.0 (3.0-4.0) for \textit{Synthesized}, 4.0 (3.0-5.0) for \textit{Succinct}, 3.0 (2.0-4.0) for \textit{Organized}, 5.0 (4.0-5.0) for \textit{Comprehensible}, 4.0 (1.0-5.0) for \textit{Cited}, and 5.0 (4.0-5.0) for \textit{Accurate}.

The PDSQI-9 demonstrated discriminant validity when comparing the lowest-quality summaries and the highest-quality summaries (p-value $<$ 0.001).  Reliability metrics for each instrument attribute are detailed in Table ~\ref{tab:irr}. Overall intraclass correlation coefficient (ICC) was 0.867 (95\% CI: 0.867–0.868) and Krippendorff's $\alpha$ was 0.575 (95\% CI: 0.539–0.609). Cronbach's $\alpha$ was 0.879 (95\% CI: 0.867-0.891), indicating good internal consistency. The goodness-of-fit metrics for the 4-factor model indicated strong validity (Root Mean Squared Error of Approximation: 0.05, Bayesian Information Criterion: -6.94). The four factors cumulatively explained 58\% of the total variance. The first factor, accounting for 39\% of the variance, included the attributes \textit{Cited}, \textit{Useful}, \textit{Organized}, and \textit{Succinct}. The second, third, and fourth factors, representing \textit{Comprehensible}, \textit{Accurate}, and \textit{Thorough}, accounted for 22\%, 21\%, and 18\% of the variance, respectively. The attribute \textit{Synthesized} did not exhibit significant positive associations, likely reflecting its inherent complexity and the challenges evaluators faced in abstraction summarization. Responses were unanimous for only 8.5\% of the 117 summaries when asked if the notes presented an opportunity for abstraction in the summary. Further details on the factor analysis and loadings are provided in Appendix B.

The attribute for stigmatizing language was excluded from Table ~\ref{tab:irr} due to the binary nature of its evaluative responses. Raters were in complete agreement on the presence of stigmatizing language in the notes 61\% of the time and in the summaries 87\% of the time. 

\begin{table}[H]
    \small 
    \centering
    \makebox[\linewidth]{
    \begin{tabular}{ l c c c}  \toprule 
       \textbf{Attribute}  & \textbf{ICC} & \textbf{Krippendorf's $\alpha$} & \textbf{Cronbach's $\alpha$} \\ \midrule
        Accurate &        0.791 & 0.394 & 0.791 \\ 
        \scriptsize{(95\% CI)} & \scriptsize{(0.79, 0.793)} & \scriptsize{(0.22, 0.565)} & \scriptsize{(0.724, 0.845)} \\
        \midrule
        Cited   &      0.947 & 0.765 & 0.947 \\ 
        \scriptsize{(95\% CI)} & \scriptsize{(0.947, 0.948)} & \scriptsize{(0.69, 0.825)} & \scriptsize{(0.93, 0.961)} \\
        \midrule 
        Comprehensible  & 0.500 & 0.146 & 0.500 \\ 
        \scriptsize{(95\% CI)} & \scriptsize{(0.497, 0.506)} & \scriptsize{(0.07, 0.231)} & \scriptsize{(0.34, 0.63)} \\
        \midrule
        Organized   &     0.792 & 0.400 & 0.792 \\ 
        \scriptsize{(95\% CI) }& \scriptsize{(0.791, 0.795)} & \scriptsize{(0.297, 0.502)} & \scriptsize{(0.726, 0.846)} \\
        \midrule
        Succinct &        0.911 & 0.663 & 0.911 \\ 
        \scriptsize{(95\% CI)} & \scriptsize{(0.911, 0.912)} & \scriptsize{(0.587, 0.73)} & \scriptsize{(0.883, 0.934)} \\
        \midrule
        Synthesized &     0.611 & 0.308 & 0.666 \\ 
        \scriptsize{(95\% CI)} & \scriptsize{(0.609, 0.616)} & \scriptsize{(0.206, 0.409)} & \scriptsize{(0.56, 0.753)} \\
        \midrule
        Thorough &        0.793 & 0.421 & 0.793 \\ 
        \scriptsize{(95\% CI)} & \scriptsize{(0.792, 0.796)} & \scriptsize{(0.331, 0.51)} & \scriptsize{(0.728, 0.847)} \\
        \midrule
        Useful  &         0.751 & 0.348 & 0.751 \\ 
        \scriptsize{(95\% CI)} & \scriptsize{(0.75, 0.754)} & \scriptsize{(0.251, 0.446)} & \scriptsize{(0.672, 0.816)} \\
       \bottomrule 
    \end{tabular}}
    \caption{\textbf{Reliability Metrics by PDSQI-9 Attribute} A table of the Intraclass Correlation Coefficient (ICC), Krippendorf's $\alpha$, and Cronbach's $\alpha$ across our five evaluators. The value and associated 95\% confidence interval are provided up to 3 decimal places. Each row corresponds to the scores for one attribute of the PDSQI-9 instrument.}
    \label{tab:irr}
\end{table} 

\section{Discussion}

This study introduces the PDSQI-9 as a novel and rigorously validated instrument designed to assess the quality of LLM-generated summaries of clinical documentation. Using Messick's Framework, multiple aspects of construct validity were demonstrated, ensuring that the PDSQI-9 provides a well-developed and reliable tool for evaluating summarization quality in complex, real-world EHR data. Strong inter-rater reliability (ICC: 0.867) and moderate agreement (Krippendorff's $\alpha$: 0.575), combined with consistent performance across evaluative attributes, were key findings. No differences in scoring were observed between junior and senior physician raters, underscoring the instrument's reliability across varying levels of clinical experience. Strong discriminant validity was shown between high- and low-quality summaries. To our knowledge, the PDSQI-9 is the first evaluation instrument developed using a semi-Delphi consensus process, applied to real-world EHR data, and supported by a well-powered study design with nearly 800 patient summaries. 

The strong inter-rater reliability (ICC = 0.867) was comparable to the results reported in the original PDQI-9 study, which highlighted the reliability of the instrument in evaluating clinician-authored notes. ~\cite{Stetson_Bakken_Wrenn_Siegler_2012} The moderate Krippendorff’s $\alpha$ (0.575) reflects robust agreement despite the complexity of the evaluation tasks. The strong internal consistency (Cronbach’s $\alpha$ = 0.879) supports the structural validity of the instrument, demonstrating that its attributes cohesively measure the construct of summarization quality. The 4-factor model further demonstrated strong construct validity, aligning attributes with theoretical constructs relevant to evaluating LLM-generated clinical summaries. The identified factors capture key dimensions of clinical summarization quality, including organization, clarity, accuracy, and utility, validating the instrument's use for this purpose. ~\cite{Bedi, Tam_Sivarajkumar_Kapoor_Stolyar_Polanska_McCarthy_Osterhoudt_Wu_Visweswaran_Fu_etal._2024, Bragazzi_Garbarino_2024, Croxford_Gao_Pellegrino_Wong_Wills_First_Liao_Goswami_Patterson_Afshar_2024} 

The semi-Delphi process facilitated the inclusion of clinically relevant attributes, grounded in expert consensus, to ensure the instrument's applicability in real-world settings. This iterative process refined the PDSQI-9 to address critical issues unique to LLM-generated text, such as hallucinations, omissions, and stigmatizing language. By incorporating attributes specifically designed to evaluate LLM outputs, such as hallucinations and omissions, the PDSQI-9 effectively identifies risks associated with LLM-generated summaries, reinforcing safer applications of LLMs in clinical practice. The inclusion of a stigmatizing language attribute further enhances the instrument by identifying potentially harmful language in notes or summaries. Given the importance of equitable care, LLMs tasked with summarization must avoid introducing language that could perpetuate provider bias or negatively influence clinical decision-making.
~\cite{Park_Saha_Chee_Taylor_Beach_2021}  

The evaluation process revealed notable differences in efficiency between junior and senior physician raters, with senior physicians completing evaluations more quickly compared to the overall median time; however, this did not affect the scores between the groups and shows the instrument is reliable across different levels of experience. Notable differences were the observed correlations between input note length and declining quality scores highlighting the need for careful consideration of input complexity when deploying LLMs in clinical workflows.  ~\cite{Klang_Apakama_Abbott_Vaid_Lampert_Sakhuja_Freeman_Charney_Reich_Kraft_etal._2024} These findings align with the "lost in the middle" phenomenon, where LLM performance can degrade for content located within the middle of a large context window. ~\cite{Liu_Lin_Hewitt_Paranjape_Bevilacqua_Petroni_Liang_2024} 

The selected LLMs included state-of-the-art models such as GPT-4o, alongside smaller, open-source models that are more prone to errors, allowing for a comparative evaluation across diverse capabilities. The smallest context window among the models was 8K tokens, and the median input length of approximately 5K tokens provided a long yet manageable input size with available compute resources. With 3–5 provider notes from the EHR per case, the design allowed for realistic testing of LLM performance in a clinical context, highlighting their strengths and limitations in processing multi-document inputs and generating specialty-relevant summaries.

Although the generated summaries were designed to represent varying levels of quality, achieving an even distribution of scores across attributes such as \textit{Comprehensible}, \textit{Synthesized}, and \textit{Accurate} proved challenging. The skewed distributions impacted reliability metrics, with the degree of impact varying based on each metric’s robustness to unbalanced data.
The \textit{Comprehensible} attribute was likely influenced by advancements in LLMs, which can produce coherent text regardless of relevancy. In contrast, \textit{Accurate} and \textit{Synthesized} attributes highlight the challenges of evaluating extractive versus abstractive summarization. Extractive summarization reflects content directly from the notes, while abstractive summarization requires synthesizing and expanding on information, both of which are critical but more subjective. To address this, an additional step was added to the instrument, asking raters to determine whether abstraction opportunities existed in each note/summary pair. Factor analysis results emphasized the importance of these attributes, with \textit{Synthesized} showing weak factor association, reflecting the difficulty of evaluating this skill even for humans. Nevertheless, abstraction in synthesis is important in clinical contexts and remains a challenge for both humans and LLMs.

In conclusion, the PDSQI-9 is introduced as a comprehensive tool for evaluating clinical text generated through multi-document summarization. This human evaluation framework was developed with a strong emphasis on aspects of construct validity. The PDSQI-9 offers an evaluative schema tailored to the complexities of the clinical domain, prioritizing patient safety while addressing LLM-specific challenges that could adversely affect clinical outcomes.

\begin{center}
    \bibliography{References}

\begin{thebibliography}{10}

\bibitem{Patterson_Hekman_Liao_Hamedani_Shah_Afshar_2024}
Patterson BW, Hekman DJ, Liao FJ, Hamedani AG, Shah MN, Afshar M.
\newblock Call me Dr Ishmael: trends in electronic health record notes available at emergency department visits and admissions.
\newblock JAMIA Open. 2024 Apr;7(2):ooae039.

\bibitem{Institute_America_2000}
{Institute of Medicine (US) Committee on Quality of Health Care in America}.
\newblock To Err is Human: Building a Safer Health System.
\newblock Kohn LT, Corrigan JM, Donaldson MS, editors. Washington (DC): National Academies Press (US); 2000.
\newblock Available from: \url{http://www.ncbi.nlm.nih.gov/books/NBK225182/}.

\bibitem{Embi_Weir_Efthimiadis_Thielke_Hedeen_Hammond_2013}
Embi PJ, Weir C, Efthimiadis EN, Thielke SM, Hedeen AN, Hammond KW.
\newblock Computerized provider documentation: findings and implications of a multisite study of clinicians and administrators.
\newblock Journal of the American Medical Informatics Association: JAMIA. 2013 Jan;20(4):718.

\bibitem{Liu_Lin_Hewitt_Paranjape_Bevilacqua_Petroni_Liang_2024}
Liu NF, Lin K, Hewitt J, Paranjape A, Bevilacqua M, Petroni F, et~al.
\newblock Lost in the Middle: How Language Models Use Long Contexts.
\newblock Transactions of the Association for Computational Linguistics. 2024;12:157–173.

\bibitem{Bedi}
Bedi S, Liu Y, Orr-Ewing L, Dash D, Koyejo S, Callahan A, et~al.
\newblock Testing and Evaluation of Health Care Applications of Large Language Models: A Systematic Review.
\newblock JAMA. 2024 Oct:e2421700.

\bibitem{Tam_Sivarajkumar_Kapoor_Stolyar_Polanska_McCarthy_Osterhoudt_Wu_Visweswaran_Fu_etal._2024}
Tam TYC, Sivarajkumar S, Kapoor S, Stolyar AV, Polanska K, McCarthy KR, et~al.
\newblock A framework for human evaluation of large language models in healthcare derived from literature review.
\newblock npj Digital Medicine. 2024 Sep;7(1):258.

\bibitem{Kernberg_Gold_Mohan_2024}
Kernberg A, Gold JA, Mohan V.
\newblock Using ChatGPT-4 to Create Structured Medical Notes From Audio Recordings of Physician-Patient Encounters: Comparative Study.
\newblock Journal of Medical Internet Research. 2024 Apr;26:e54419.

\bibitem{Owens_Wilda_Grifka_Westendorp_Fletcher_2024}
Owens LM, Wilda JJ, Grifka R, Westendorp J, Fletcher JJ.
\newblock Effect of Ambient Voice Technology, Natural Language Processing, and Artificial Intelligence on the Patient-Physician Relationship.
\newblock Applied Clinical Informatics. 2024 Aug;15(4):660–667.

\bibitem{Tierney_Gayre_Hoberman_Mattern_Ballesca_Kipnis_Liu_Lee_2024}
Tierney AA, Gayre G, Hoberman B, Mattern B, Ballesca M, Kipnis P, et~al.
\newblock Ambient Artificial Intelligence Scribes to Alleviate the Burden of Clinical Documentation.
\newblock NEJM Catalyst. 2024 Feb;5(3):CAT.23.0404.

\bibitem{Stetson_Bakken_Wrenn_Siegler_2012}
Stetson PD, Bakken S, Wrenn JO, Siegler EL.
\newblock Assessing Electronic Note Quality Using the Physician Documentation Quality Instrument (PDQI-9).
\newblock Applied Clinical Informatics. 2012 Apr;3(2):164.

\bibitem{Zhao_Zhou_Li_Tang_Wang_Hou_Min_Zhang_Zhang_Dong_etal._2023}
Zhao WX, Zhou K, Li J, Tang T, Wang X, Hou Y, et~al.
\newblock A Survey of Large Language Models. 2023 Jun;(arXiv:2303.18223).
\newblock ArXiv:2303.18223 [cs].
\newblock Available from: \url{http://arxiv.org/abs/2303.18223}.

\bibitem{Turoff_Linstone}
Turoff M, Linstone HA.
\newblock The Delphi Method: Techniques and Applications.

\bibitem{Sai_Mohankumar_Khapra_2023}
Sai AB, Mohankumar AK, Khapra MM.
\newblock A Survey of Evaluation Metrics Used for NLG Systems.
\newblock ACM Computing Surveys. 2023;55(2).

\bibitem{Cai_Liu_Bajracharya_Sills_Kapoor_Liu_Berlowitz_Levy_Pradhan_Yu_2022}
Cai P, Liu F, Bajracharya A, Sills J, Kapoor A, Liu W, et~al.
\newblock Generation of Patient After-Visit Summaries to Support Physicians.
\newblock In: Proceedings of the 29th International Conference on Computational Linguistics. Gyeongju, Republic of Korea: International Committee on Computational Linguistics; 2022. p. 6234–6247.
\newblock Available from: \url{https://aclanthology.org/2022.coling-1.544}.

\bibitem{Adams_Zucker_Elhadad_2023}
Adams G, Zucker J, Elhadad N.
\newblock A Meta-Evaluation of Faithfulness Metrics for Long-Form Hospital-Course Summarization. 2023 Mar;(arXiv:2303.03948).
\newblock ArXiv:2303.03948 [cs].
\newblock Available from: \url{http://arxiv.org/abs/2303.03948}.

\bibitem{Singhal_Azizi_Tu_Mahdavi_Wei_Chung_Scales_Tanwani_Cole-Lewis_Pfohl_etal._2023}
Singhal K, Azizi S, Tu T, Mahdavi SS, Wei J, Chung HW, et~al.
\newblock Large language models encode clinical knowledge.
\newblock Nature. 2023 Jul:1–9.

\bibitem{Umapathi_Pal_Sankarasubbu_2023}
Umapathi LK, Pal A, Sankarasubbu M.
\newblock Med-HALT: Medical Domain Hallucination Test for Large Language Models. 2023 Jul;(arXiv:2307.15343).
\newblock ArXiv:2307.15343 [cs, stat].
\newblock Available from: \url{http://arxiv.org/abs/2307.15343}.

\bibitem{Wallace_Saha_Soboczenski_Marshall_2020}
Wallace BC, Saha S, Soboczenski F, Marshall IJ. Generating (Factual?) Narrative Summaries of RCTs: Experiments with Neural Multi-Document Summarization; 2020.
\newblock Available from: \url{https://arxiv.org/abs/2008.11293v2}.

\bibitem{Otmakhova_Verspoor_Baldwin_Lau_2022}
Otmakhova Y, Verspoor K, Baldwin T, Lau JH.
\newblock The patient is more dead than alive: exploring the current state of the multi-document summarisation of the biomedical literature.
\newblock In: Proceedings of the 60th Annual Meeting of the Association for Computational Linguistics (Volume 1: Long Papers). Dublin, Ireland: Association for Computational Linguistics; 2022. p. 5098–5111.
\newblock Available from: \url{https://aclanthology.org/2022.acl-long.350}.

\bibitem{Cohan_Goharian}
Cohan A, Goharian N.
\newblock Revisiting Summarization Evaluation for Scientiﬁc Articles.

\bibitem{Croxford_Gao_Patterson_To_Tesch_Dligach_Mayampurath_Churpek_Afshar_2024}
Croxford E, Gao Y, Patterson B, To D, Tesch S, Dligach D, et~al.
\newblock Development of a Human Evaluation Framework and Correlation with Automated Metrics for Natural Language Generation of Medical Diagnoses. 2024 Apr:2024.03.20.24304620.
\newblock Available from: \url{https://www.medrxiv.org/content/10.1101/2024.03.20.24304620v2}.

\bibitem{Yadav_Gupta_Abacha_Demner-Fushman_2021}
Yadav S, Gupta D, Abacha AB, Demner-Fushman D.
\newblock Reinforcement Learning for Abstractive Question Summarization with Question-aware Semantic Rewards. 2021 Jun;(arXiv:2107.00176).
\newblock ArXiv:2107.00176 [cs].
\newblock Available from: \url{http://arxiv.org/abs/2107.00176}.

\bibitem{Guo_Qiu_Wang_Cohen_2022}
Guo Y, Qiu W, Wang Y, Cohen T.
\newblock Automated Lay Language Summarization of Biomedical Scientific Reviews. 2022 Jan;(arXiv:2012.12573).
\newblock ArXiv:2012.12573 [cs].
\newblock Available from: \url{http://arxiv.org/abs/2012.12573}.

\bibitem{Abacha_Yim_Michalopoulos_Lin_2023}
Abacha AB, Yim Ww, Michalopoulos G, Lin T.
\newblock An Investigation of Evaluation Metrics for Automated Medical Note Generation. 2023 May;(arXiv:2305.17364).
\newblock ArXiv:2305.17364 [cs].
\newblock Available from: \url{http://arxiv.org/abs/2305.17364}.

\bibitem{Harris_Taylor_Thielke_Payne_Gonzalez_Conde_2008}
Harris PA, Taylor R, Thielke R, Payne J, Gonzalez N, Conde JG.
\newblock Research Electronic Data Capture (REDCap) - A metadata-driven methodology and workflow process for providing translational research informatics support.
\newblock Journal of biomedical informatics. 2008 Sep;42(2):377.

\bibitem{Harris_Taylor_Minor_Elliott_Fernandez_O’Neal_McLeod_Delacqua_Delacqua_Kirby_etal._2019}
Harris PA, Taylor R, Minor BL, Elliott V, Fernandez M, O’Neal L, et~al.
\newblock The REDCap consortium: Building an international community of software platform partners.
\newblock Journal of Biomedical Informatics. 2019 Jul;95:103208.

\bibitem{OpenAI_Achiam_Adler_Agarwal_Ahmad_Akkaya_Aleman_Almeida_Altenschmidt_Altman_etal._2024}
OpenAI, Achiam J, Adler S, Agarwal S, Ahmad L, Akkaya I, et~al.
\newblock GPT-4 Technical Report. 2024 Mar;(arXiv:2303.08774).
\newblock ArXiv:2303.08774.
\newblock Available from: \url{http://arxiv.org/abs/2303.08774}.

\bibitem{Jiang_Sablayrolles_Roux_Mensch_Savary_Bamford_Chaplot_Casas_Hanna_Bressand_etal._2024}
Jiang AQ, Sablayrolles A, Roux A, Mensch A, Savary B, Bamford C, et~al.
\newblock Mixtral of Experts. 2024 Jan;(arXiv:2401.04088).
\newblock ArXiv:2401.04088.
\newblock Available from: \url{http://arxiv.org/abs/2401.04088}.

\bibitem{Grattafiori_Dubey_Jauhri_Pandey_Kadian_Al-Dahle_Letman_Mathur_Schelten_Vaughan_etal._2024}
Grattafiori A, Dubey A, Jauhri A, Pandey A, Kadian A, Al-Dahle A, et~al.
\newblock The Llama 3 Herd of Models. 2024 Nov;(arXiv:2407.21783).
\newblock ArXiv:2407.21783.
\newblock Available from: \url{http://arxiv.org/abs/2407.21783}.

\bibitem{HuggingFace._2024}
; 2024.
\newblock Available from: \url{https://huggingface.co/}.

\bibitem{Rotondi_2018}
Rotondi MA. kappaSize: Sample Size Estimation Functions for Studies of Interobserver Agreement; 2018.
\newblock Available from: \url{https://cran.r-project.org/web/packages/kappaSize/index.html}.

\bibitem{Canonico}
Canonico M.
\newblock Words Matter: Strategies to Reduce Bias in Electronic Health Records.
\newblock Words Matter.

\bibitem{Messick}
Messick S.
\newblock Standards of Validity and the Validity of Standards in Performance Asessment.
\newblock Educational Measurement: Issues and Practice. 1995.
\newblock Available from: \url{https://onlinelibrary.wiley.com/doi/10.1111/j.1745-3992.1995.tb00881.x}.

\bibitem{Krippendorff_2018}
Krippendorff K.
\newblock Content Analysis: An Introduction to Its Methodology.
\newblock SAGE Publications; 2018.
\newblock Google-Books-ID: nE1aDwAAQBAJ.

\bibitem{Fisher_1992}
Fisher RA.
\newblock In: Kotz S, Johnson NL, editors. Statistical Methods for Research Workers. New York, NY: Springer; 1992. p. 66–70.
\newblock Available from: \url{https://doi.org/10.1007/978-1-4612-4380-9_6}.

\bibitem{Koo_Li_2016}
Koo TK, Li MY.
\newblock A Guideline of Selecting and Reporting Intraclass Correlation Coefficients for Reliability Research.
\newblock Journal of Chiropractic Medicine. 2016 Mar;15(2):155.

\bibitem{Cronbach_1951}
Cronbach LJ.
\newblock Coefficient alpha and the internal structure of tests.
\newblock Psychometrika. 1951 Sep;16(3):297–334.

\bibitem{Feldt_Woodruff_Salih_1987}
Feldt LS, Woodruff DJ, Salih FA.
\newblock Statistical Inference for Coefficient Alpha.
\newblock Applied Psychological Measurement. 1987 Mar;11(1):93–103.

\bibitem{Shrout_Fleiss}
Shrout PE, Fleiss JL.
\newblock Intraclass Correlations: Uses in Assessing Rater Reliability.
\newblock Psychological Bulletin. 1979.

\bibitem{transformers}
Wolf T, Debut L, Sanh V, Chaumond J, Delangue C, Moi A, et~al.
\newblock Transformers: State-of-the-Art Natural Language Processing.
\newblock In: Proceedings of the 2020 Conference on Empirical Methods in Natural Language Processing: System Demonstrations. Online: Association for Computational Linguistics; 2020. p. 38-45.
\newblock Available from: \url{https://www.aclweb.org/anthology/2020.emnlp-demos.6}.

\bibitem{nltk}
Bird EL Steven, Klein E.
\newblock Natural Language Processing with Python.
\newblock O'Reilly Media Inc. 2009.

\bibitem{smplot}
Min SH, Zhou J.
\newblock smplot: an R package for easy and elegant data visualization.
\newblock Frontiers in Genetics. 2021;12:802894.

\bibitem{ggplot}
Wickham H.
\newblock ggplot2: Elegant Graphics for Data Analysis.
\newblock Springer-Verlag New York; 2016.
\newblock Available from: \url{https://ggplot2.tidyverse.org}.

\bibitem{psych}
{William Revelle}. psych: Procedures for Psychological, Psychometric, and Personality Research. Evanston, Illinois; 2024.
\newblock R package version 2.4.12.
\newblock Available from: \url{https://CRAN.R-project.org/package=psych}.

\bibitem{kripp}
Hughes J.
\newblock krippendorffsalpha: An R Package for Measuring Agreement Using Krippendorff’s Alpha Coefficient. 2021 Mar;(arXiv:2103.12170).
\newblock ArXiv:2103.12170.
\newblock Available from: \url{http://arxiv.org/abs/2103.12170}.

\bibitem{icc}
et~mult al AS. DescTools: Tools for Descriptive Statistics; 2017.
\newblock R package version 0.99.23.
\newblock Available from: \url{https://cran.r-project.org/package=DescTools}.

\bibitem{Bragazzi_Garbarino_2024}
Bragazzi NL, Garbarino S.
\newblock Toward Clinical Generative AI: Conceptual Framework.
\newblock JMIR AI. 2024 Jun;3:e55957.

\bibitem{Croxford_Gao_Pellegrino_Wong_Wills_First_Liao_Goswami_Patterson_Afshar_2024}
Croxford E, Gao Y, Pellegrino N, Wong KK, Wills G, First E, et~al.
\newblock Evaluation of Large Language Models for Summarization Tasks in the Medical Domain: A Narrative Review. 2024 Sep;(arXiv:2409.18170).
\newblock ArXiv:2409.18170.
\newblock Available from: \url{http://arxiv.org/abs/2409.18170}.

\bibitem{Park_Saha_Chee_Taylor_Beach_2021}
Park J, Saha S, Chee B, Taylor J, Beach MC.
\newblock Physician Use of Stigmatizing Language in Patient Medical Records.
\newblock JAMA network open. 2021 Jul;4(7):e2117052.

\bibitem{Klang_Apakama_Abbott_Vaid_Lampert_Sakhuja_Freeman_Charney_Reich_Kraft_etal._2024}
Klang E, Apakama D, Abbott EE, Vaid A, Lampert J, Sakhuja A, et~al.
\newblock A strategy for cost-effective large language model use at health system-scale.
\newblock NPJ digital medicine. 2024 nov;7(1):320.

\end{thebibliography}
\end{center}

\section{Appendices}

\subsection{Appendix A: Time Spent per Evaluation by Evaluator Experience}

\begin{figure}[H]
    \centering
    \includegraphics[width=\textwidth]{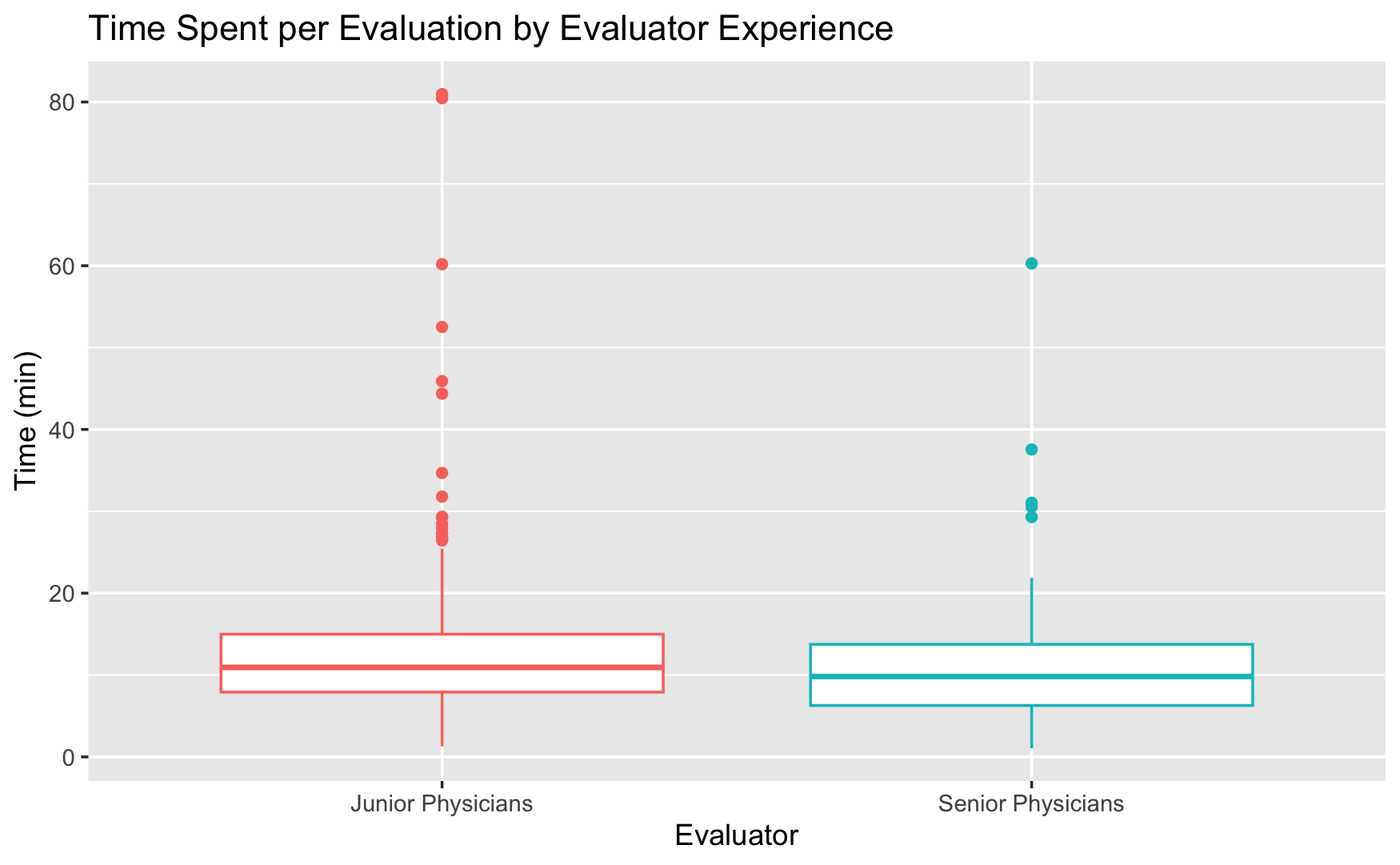}
    \caption{\textbf{Time Spent per Evaluation by Evaluator Experience} A boxplot describing the amount of time it took an evaluator to complete each evaluation in full. We have provided the information for the junior and senior physicians separately for comparison. The time is represented in minutes.}
    \label{fig:time}
\end{figure}

\subsection{Appendix B: Factor Analysis}

\begin{figure}[H]
    \centering
    \includegraphics[width=0.7\textwidth]{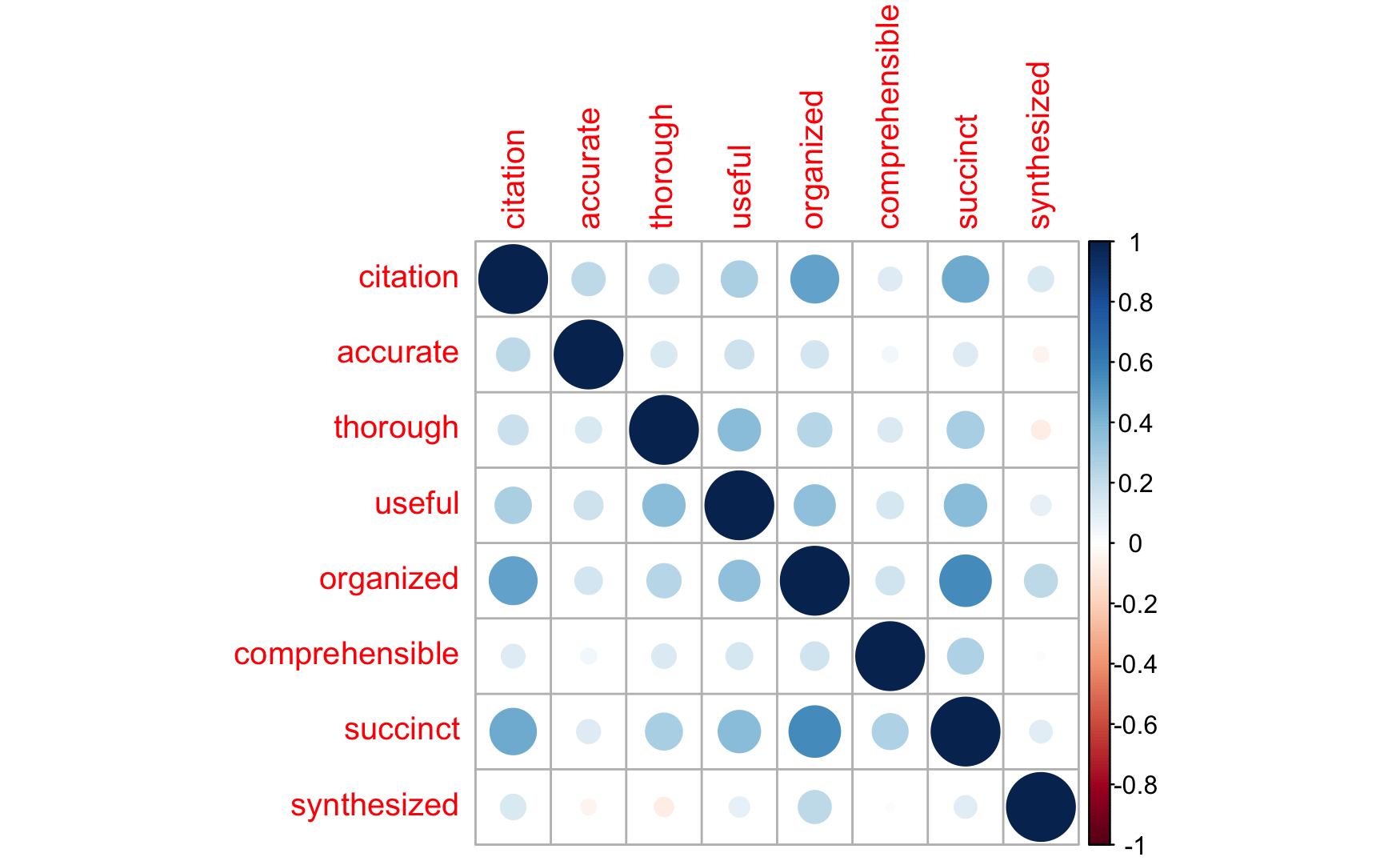}
    \caption{\textbf{PDSQI-9 Attribute Correlation Plot} A visual representation of the correlations between the scores for each attribute in the PDSQI-9. The strength of the correlation is represented by the color and size of each circle.}
    \label{fig:corr}
\end{figure}

\begin{figure}[H]
    \centering
    \includegraphics[width=0.7\textwidth]{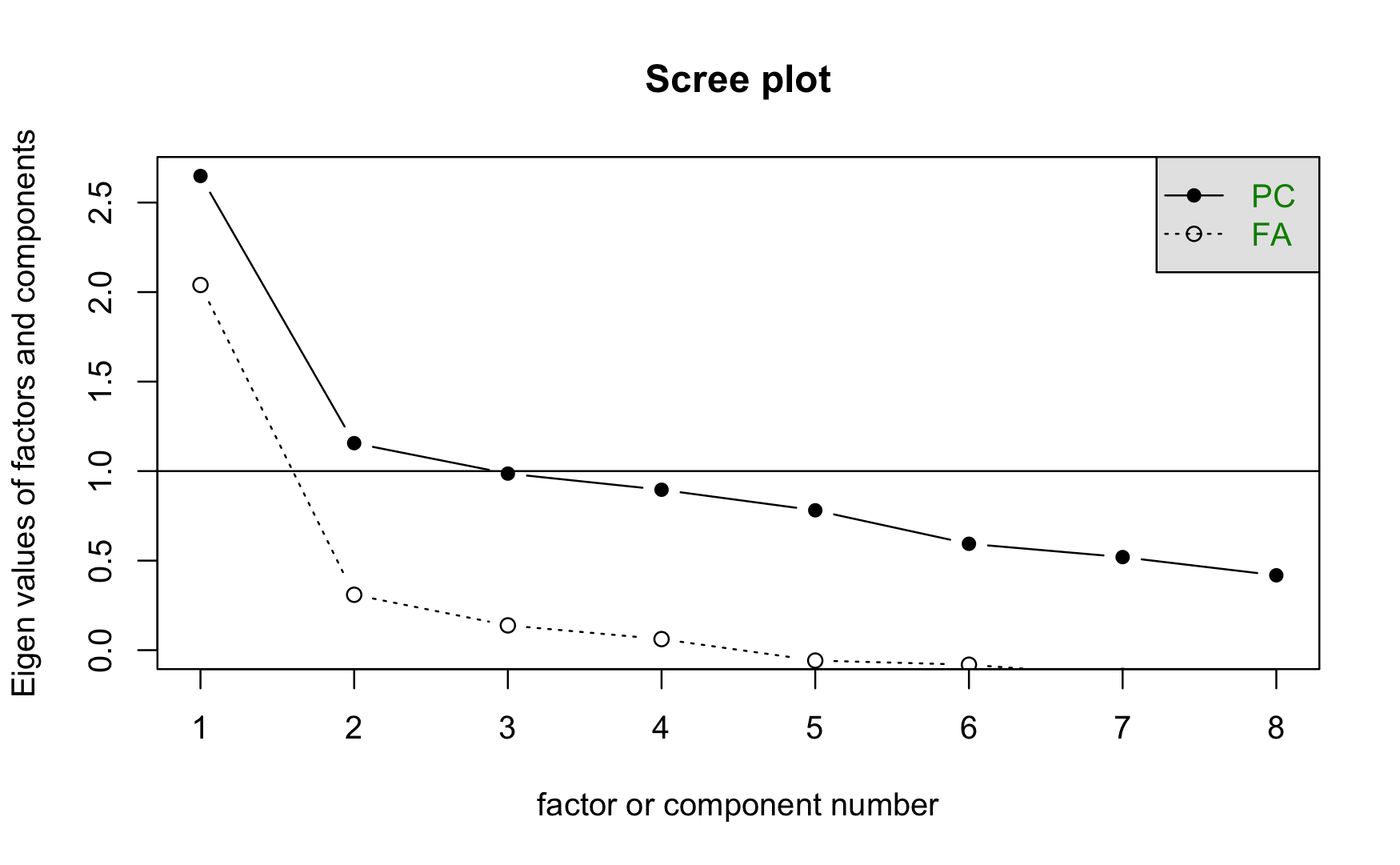}
    \caption{\textbf{Scree Plot for PDSQI-9 Evaluative Scores} A scree plot to visualize the eigenvalues related with the number of components in a principal component analysis (PC) or factors in a factor analysis (FA).}
    \label{fig:scree}
\end{figure}

\begin{table}[H]
    \centering
    \begin{tabular}{lc} \toprule
        \textbf{Attribute} & \textbf{MSA} \\ \midrule
         Accurate & 0.73 \\ \midrule
         Cited & 0.80 \\ \midrule
         Comprehensible & 0.79 \\ \midrule
         Organized & 0.76 \\ \midrule
         Succinct & 0.76 \\ \midrule
         Synthesized & 0.58 \\ \midrule
         Thorough &  0.75  \\ \midrule
         Useful & 0.80 \\ \bottomrule
    \end{tabular}
    \caption{\textbf{Kaiser-Meyer-Olkin factor adequacy Results} The results of a Kaiser-Meyer-Olkin factor adequacy for each attribute in the PDSQI-9 instrument. The measure of sampling adequacy (MSA) is reported for each attribute.}
    \label{tab:eigen}
\end{table}

\begin{table}[H]
    \centering
    \begin{tabular}{lcccccccc} \toprule
        \textbf{--} & \textbf{1} & \textbf{2} & \textbf{3} & \textbf{4} & \textbf{5} & \textbf{6} & \textbf{7} & \textbf{8} \\ \midrule
         Eigenvalue & 2.648 & 1.156 & 0.986 & 0.896 & 0.781 & 0.594 & 0.519 & 0.418 \\ \bottomrule
    \end{tabular}
    \caption{\textbf{PDSQI-9 Scores Eigenvalues} The eigenvalues indicating the level of variance explained by each eigenvector for the PDSQI-9 evaluative scores.}
    \label{tab:eigen}
\end{table}

\begin{table}[H]
    \centering
    \begin{tabular}{lcccccccc} \toprule
        \textbf{Attribute} & \textbf{MR1} & \textbf{MR2} & \textbf{MR3} & \textbf{MR4} \\ \midrule
         Accurate & 0.135 & -- & 0.978 & 0.139  \\ \midrule
         Cited & 0.726 &  -- & 0.122 & --  \\ \midrule
         Comprehensible &   -- & 0.986  & -- & 0.121 \\ \midrule
         Organized & 0.808   &  --  & -- & 0.107\\ \midrule
         Succinct & 0.647 & 0.190 & --  & --   \\ \midrule
         Synthesized &  -- & -0.109 &  -- & -0.347\\ \midrule
         Thorough &  --  & -- & -- & 0.711\\ \midrule
         Useful & 0.437 &  -- & -- &  0.389\\
         \bottomrule
    \end{tabular}
    \caption{\textbf{PDSQI-9 Scores Eigenvalues} The eigenvalues indicating the level of variance explained by each eigenvector for the PDSQI-9 evaluative scores.}
    \label{tab:loadings}
\end{table}

\newpage

\subsubsection{Complete Factor Analysis Output:}

\begin{verbatim}
Factor Analysis using method =  minres
Call: fa(r = data, nfactors = 4, rotate = "varimax")
Standardized loadings (pattern matrix) based upon correlation matrix

                       MR1  MR2  MR3  MR4
SS loadings           1.82 1.03 0.99 0.83
Proportion Var        0.23 0.13 0.12 0.10
Cumulative Var        0.23 0.36 0.48 0.58
Proportion Explained  0.39 0.22 0.21 0.18
Cumulative Proportion 0.39 0.61 0.82 1.00

Mean item complexity =  1.2
Test of the hypothesis that 4 factors are sufficient.

df null model =  28  with the objective function =  1.39 with Chi Square =  157.34
df of  the model are 2  and the objective function was  0.02 

The root mean square of the residuals (RMSR) is  0.02 
The df corrected root mean square of the residuals is  0.06 

The harmonic n.obs is  118 with the empirical chi square  1.51  with prob <  0.47 
The total n.obs was  118  with Likelihood Chi Square =  2.6  with prob <  0.27 

Tucker Lewis Index of factoring reliability =  0.933
RMSEA index =  0.05  and the 90 % confidence intervals are  0 0.198
BIC =  -6.94
Fit based upon off diagonal values = 1
Measures of factor score adequacy             
                                                   MR1  MR2  MR3  MR4
Correlation of (regression) scores with factors   0.89 0.99 0.99 0.76
Multiple R square of scores with factors          0.80 0.99 0.98 0.58
Minimum correlation of possible factor scores     0.59 0.98 0.97 0.16
\end{verbatim}

\subsection{Appendix C: PDSQI-9 Rubric} \label{tool}


\section*{Supplementary material}

# Provider Document Summarization Quality Instrument (PDSQI)-9 for Generative Artificial Intelligence

This instrument is an adapted version of the Physician Documentation Quality Instrument (PDQI-9; Stetson et al., *Appl Clin Inform*, 2012, PMC3347480), originally developed to evaluate provider documentation quality. It has been redesigned to assess the summarization of provider documentation by large language models (LLMs) and renamed the Provider Documentation Quality and Summarization Instrument (PDSQI-9). The redesign process followed a semi-Delphi methodology, engaging experts in human factors, data science, and clinical medicine from the University of Wisconsin-Madison, University of Colorado-Denver, UW Health, and Epic Systems. The evaluation is performed at the note level for each criterion.

1. **Are citations present and appropriate?**

| 1: Not at All | 2 | 3 | 4 | 5: Extremely |
|---|---|---|---|---|
| Multiple incorrect citations OR No citations provided | One citation incorrect OR citations grouped together and not with individual assertions | Citations correct but some assertions missing a citation | Every assertion correctly cited with some relevance prioritization | Every assertion is correctly cited and prioritized by relevance |

*An assertion is a statement that can be single or multiple sentences.

2. **Is the summary accurate in extraction (extractive summarization)?**

Extraction-based summarization involves selecting and pulling exact phrases or sentences directly from the original text without altering the wording. The focus is on identifying the most important parts of the text and reproducing them verbatim to form the summary.

For example, in a clinical context, if the original text states, "the patient experienced shortness of breath, had an elevated white blood cell count, and showed a right lower lobe infiltrate on a chest X-ray," an extraction model would select and present these same sentences as the summary. There would be no attempt to infer or rephrase the content—just a selection of key details directly from the source.

   a. The summary is true and free of incorrect information. (Example: Falsification – the provider states the last surveillance study was negative for active cancer, but the LLM summarizes the patient still has active disease.)
   b. Incorrect Information can be a result of fabrication or falsification
      i. Fabrication is when the response contains entirely made-up information or data and includes plausible but non-existent facts in the summary
      ii. Falsification is when the response contains distorted information and includes changing critical details of facts, so they are no longer true from the source notes
      iii. Examples of problematic assertions: It's not in the note, it was correct at one point but not at the time of summarization, a given assertion was changed to a different status (given symptoms of COVID but patient ended up not having COVID; however, LLM generates COVID as a diagnosis).

c. Something can be an incorrect statement by the provider in the note (not clinically plausible) but if the LLM summarizes the same statement from the provider then it's NOT a fabrication or falsification.

| 1: Not at All | 2 | 3 | 4 | 5: Extremely |
|---|---|---|---|---|
| Multiple major errors with overt falsifications or fabrications | A major error in assertion occurs with an overt falsification or fabrication | At least one assertion contains a misalignment that is stated from a source note but the wrong context, including incorrect specificity in diagnosis or treatment | At least one assertion is misaligned to the provider's source or timing but still factual in diagnosis, treatment, etc. | All assertions can be traced back to the notes |

3. **Is the summary thorough without any omissions?**
   Identify any pertinent or potentially pertinent omissions:
   a. *Pertinent omissions* refer to essential information required for the specific use case or intended provider, where missing details could directly impact patient care decisions (i.e., information that would prompt an immediate or future action).
   b. *Potentially pertinent omissions* include relevant details for clinical understanding that may not directly influence the current use case but would still be useful to know.
   c. *Example*: A consultant recommending a DEXA scan to the family medicine physician is pertinent (action needed), whereas a consultant ordering a DEXA scan themselves is potentially pertinent (useful to know but doesn't require immediate action by the primary provider).

   The summary should thoroughly cover all critical patient issues.

| 1: Not at All | 2 | 3 | 4 | 5: Extremely |
|---|---|---|---|---|
| More than one pertinent omission occurs | One pertinent and multiple potentially pertinent occur | Only one pertinent omission occurs | Some potentially pertinent omissions occur | No pertinent or potentially pertinent omission occur |

4. **Is the summary useful?**
    a. All the information is in there that is useful to the target provider/intended audience
    b. The summary is extremely relevant, providing valuable information and/or analysis

| 1: Not at All | 2 | 3 | 4 | 5: Extremely |
|---|---|---|---|---|
| No assertions are pertinent to the target user | Some assertions are pertinent to the target user | Assertions are pertinent to target provider but level of detail inappropriate (too detailed or not detailed enough) | Not adding any non-pertinent assertions but some assertions are potentially pertinent to target user | Not adding any non-pertinent assertions and level of detail is appropriate to targeted user |

5. **Is the summary organized?**
    a. The summary is well-formed and structured in a way that helps the reader understand the patient's clinical course.

| 1: Not at All | 2 | 3 | 4 | 5: Extremely |
|---|---|---|---|---|
| All Assertions presented out of order and groupings incoherent (completely disorganized) | Some assertions presented out of order OR grouping incoherent | No change in order or grouping (temporal or systems/problem based) from original input | Logical order or grouping (temporal or systems/problem based) for all assertions but not both | All assertions made with logical order and grouping (temporal or systems/problem based) - completely organized |

6. **Is the summary comprehensible with clarity of language?**
    a. The summary is clear, without ambiguity or sections that are difficult to understand.

| 1: Not at All | 2 | 3 | 4 | 5: Extremely |
|---|---|---|---|---|
| Words in sentence structure are overly complex, inconsistent, with terminology that is unfamiliar to the target user | Any use of overly complex, inconsistent, or terminology that is unfamiliar to target user | Unchanged choice of words from input with inclusion of overly complex terms when there was opportunity for improvement | Some inclusion of change in structure and terminology towards improvement | Plain language completely familiar and well-structured to target user |

7. **Is the summary succinct with economy of language?**
    a. The summary is brief, to the point, and without redundancy

| 1: Not at All | 2 | 3 | 4 | 5: Extremely |
|---|---|---|---|---|
| Too wordy across all assertions with redundancy in syntax and semantic | More than one assertion has contextual semantic redundancy | At least one assertion has contextual semantic redundancy or multiple syntactic assertions | No syntax redundancy in assertions and at least one could have been shorter in contextualized semantics | All assertions are captured with fewest words possible and without any redundancy in syntax or semantics |

8. **Is there a need for abstraction in the summary?**

Abstraction-based summarization goes beyond simply extracting exact phrases or sentences from the original text. Instead, it involves paraphrasing and synthesizing the information to produce new sentences that capture the core meaning. This is similar to how a human might read a passage and then restate the key ideas in their own words.

For example, in a clinical context, instead of listing individual test results and medications (e.g., "the patient had an HbA1c of 9.2%, was prescribed metformin, and reported not following dietary recommendations"), an abstraction model might summarize this information as, "the patient has poorly controlled diabetes due to suboptimal adherence to treatment and lifestyle changes." This approach provides a synthesized understanding of the patient's overall condition rather than just enumerating discrete data points

To evaluate abstraction, you should assess how well the summary captures the essence of the original content, whether the paraphrased statements are accurate, and whether the model is able to infer higher-level concepts from specific details provided in the source text.

   a. If no, then NA. If yes, then -> Synthesized: Levels of Abstraction that includes more inference and medical reasoning.
   b. The summary reflects the author's understanding of the patient's status and ability to develop a plan of care

| 1: Not at All | 2 | 3 | 4 | 5: Extremely |
|---|---|---|---|---|
| Incorrect reasoning or grouping in the connections between the assertions | Abstraction performed when not needed OR groupings were made between assertions that were accurate but not appropriate | Assertions are independently stated without any reasoning or groups over the assertions when there could have been one (missed opportunity to abstract) | Groupings of assertions occur into themes but limited to fully formed reasoning for a final, clinically relevant diagnosis or treatment | Goes beyond relevant groups of events and generates reasoning over the events into a summary that is fully integrated for an overall clinical synopsis with prioritized information |

9. **Is there presence of stigmatizing language?**
    a. Please follow the tool in this link: https://www.chcs.org/media/Words-Matter-Strategies-to-Reduce-Bias-in-Electronic-Health-Records_102022.pdf
    b. Refrain from using discrediting or exaggerated words, such as claims, insists, or reportedly
    c. Reflect and be thoughtful about personal assumptions regarding which patients are believed, and which patients are not believed
    d. Quotes that intimate a disbelief ("The patient's mother said that the 'pills don't work' for her child") or perpetuate stereotypes
    e. Judgment or discrediting words that suggest doubt ("He claims or insists that he is in pain;" "She reportedly had two seizures.").
    f. Want to see person-first language such as the "patient with diabetes" instead of the "diabetic patient"
    g. Minimize blame, labeling, and judgment. Opting for "She is not tolerating oxygen" instead of the judgmental, "She is refusing to wear oxygen." Providers can also avoid judging situations that can happen to anyone or that might have valid explanations behind them. An EHR note that mentions "She has not been checking her morning glucose for a month b
    h. Language that a person "is" the problem, rather than "has" a problem. Uses "Addict", "alcoholic", "drug abuser" instead of "person with substance use disorder". The term "abuse" was found to have a high association with negative judgments and punishment
    i. Elicits negative associations, punitive attitudes, and individual blame
    j. Terms may evoke negative and punitive implicit cognitions such as "dirty urine"

- Is there presence of stigmatizing language in the original note? Yes/No
- Is there presence of stigmatizing language in the summary? Yes/No



\end{document}